\documentclass[runningheads]{llncs}

\usepackage{eccv}

\usepackage{eccvabbrv}

\usepackage{graphicx}
\usepackage{booktabs}
\usepackage{multirow}
\usepackage{amsmath}
\usepackage{wrapfig}
\usepackage{array}
\usepackage{rotating} 
\usepackage[accsupp]{axessibility}  

\usepackage{hyperref}

\usepackage{orcidlink}

\usepackage{xcolor}
\usepackage[normalem]{ulem}

\definecolor{darkred}{RGB}{171,35,40}

\begin{document}

\title{Reshaping the Online Data Buffering and Organizing Mechanism for Continual Test-Time Adaptation} 

\titlerunning{Reshaping the Online Data Buffering and Organizing Mechanism for CTTA}

\author{Zhilin Zhu\inst{1,2} \and
Xiaopeng Hong\inst{1,2}\thanks{Corresponding author} \and
Zhiheng Ma\inst{3,4,5} \and
Weijun Zhuang\inst{1,2} \and
Yaohui Ma\inst{1,5} \and
Yong Dai\inst{2} \and
Yaowei Wang\inst{2,1}}

\authorrunning{Z. Zhu et al.}

\institute{Harbin Institute of Technology \and
Pengcheng Laboratory \and Shenzhen University of Advanced Technology \and Guangdong Provincial Key Laboratory of Computility Microelectronics \and Shenzhen Institute of Advanced Technology, Chinese Academy of Science\\
\email{\{zhuzhilin, weijunzhuang\}@stu.hit.edu.cn hongxiaopeng@ieee.org zh.ma@siat.ac.cn yh.ma@siat.ac.cn chd-dy@foxmail.com wangyw@pcl.ac.cn}}

\maketitle

\begin{abstract}

{Continual Test-Time Adaptation (CTTA) involves adapting a pre-trained source model to continually changing unsupervised target domains. In this paper, we systematically analyze the challenges of this task: online environment, unsupervised nature, and the risks of error accumulation and catastrophic forgetting under continual domain shifts. To address these challenges, we reshape the online data buffering and organizing mechanism for CTTA. We propose an {uncertainty-aware buffering approach} to identify {and aggregate} significant samples with high certainty from the unsupervised, single-pass data stream. {Based on this}, we propose a graph-based class relation preservation constraint to overcome catastrophic forgetting. Furthermore, a pseudo-target replay objective is used to mitigate error accumulation. Extensive experiments demonstrate the superiority of our method in both segmentation and classification CTTA tasks. Code is available at \href{https://github.com/z1358/OBAO}{https://github.com/z1358/OBAO}.}

\end{abstract}

\section{Introduction}
\label{sec:intro}
Deep learning models have exhibited remarkable advancements in scenarios characterized by a consistent data distribution between training and test environments. However, the deployment of machine perception systems in the real world frequently exposes them to non-stationary and constantly changing environments{~\cite{koh2021wilds,Tao_2020_CVPR,fan2024dynamic,wang2023isolation,yaosocialized}}.  When such phenomena occur (\eg, unexpected domain shifts over time), these models are prone to severe performance degradation~\cite{hendrycks2018benchmarking,niu2023towards,Wang_2022_CVPR}. 
Consequently, the development of techniques for continual adaptation during the unsupervised deployment phase is essential for enhancing the reliability of machine perception systems.

\begin{table}[t]
    \centering
    \caption{Comparison of different adaptation settings. We differentiate along four dimensions: the available form of the source domain data and target domain data, the distribution and access mode of the {target domain data stream} during adaptation.
    }
    \label{tab:settings}
    \resizebox{0.96\textwidth}{!}{
    \begin{tabular}{l c  c   c  c }
    \toprule
     \multirow{2}{*}{Setting} & \multicolumn{2}{c}{Data}  & \multicolumn{2}{c}{{Target domain data stream}}\\
    \cmidrule(lr){2-3}\cmidrule(lr){4-5}
     & Source & Target & Distribution & Access mode \\
    \midrule
    Fine-tuning & - & $\mathcal{X}^{T},\mathcal{Y}^{T}$ & stationary & offline (multi-epoch) \\
    Domain Generalization & $\mathcal{X}^{S},\mathcal{Y}^{S}$ & - & - & - \\
    Domain Adaptation & $\mathcal{X}^{S},\mathcal{Y}^{S}$ & $\mathcal{X}^{T}$ & stationary & offline (multi-epoch) \\
    Test-Time Training & $\mathcal{X}^{S},\mathcal{Y}^{S}$ & $\mathcal{X}^{T}$ & stationary & online (single-pass) \\
    Test-Time Adaptation & - & $\mathcal{X}^{T}$ & stationary & offline or online\\
    Continual Test-Time Adaptation  & - & $\mathcal{X}^T_1\to\cdots\to \mathcal{X}^T_n$ & continually changing & online (single-pass) \\
    \bottomrule
    \end{tabular}}
\end{table}

With sufficient source labeled data accessible, the fields of domain generalization~\cite{lee2022cross,zhou2022domain,zhang2022mvdg,li2018domain} and domain adaptation~\cite{hwang2022combating,lin2022prototype,li2021generalized,li2021transferable} have witnessed considerable success. However, a particularly challenging scenario arises when no source data is available, and unsupervised adaptation must happen at {test time} when unlabeled test inputs arrive. This task is known as test-time adaptation {(TTA)~\cite{tan2024uncertainty,choi2022improving,chen2022contrastive}}. Given the rarity of encountering {only} a singular domain shift in practical applications, the community has recently been increasingly focusing on continual test-time adaptation (CTTA)~\cite{Wang_2022_CVPR,Brahma_2023_CVPR,Dobler_2023_CVPR,ijcai2023p183,liu2024vida}, \ie, evaluating the ability of methods to adapt to continual domain shifts. We summarize {the differences between existing adaptation settings} in \cref{tab:settings}. Obviously, the considered CTTA setting is the most practical and challenging.

Within this setting, we expose the {key challenges} as follows: \textbf{(1) {Online environment.}} This denotes a scenario where data is processed in a single pass, necessitating immediate predictions from the model. Due to the characteristics of the online environment, it is difficult for the model to obtain a sufficient number of informative samples, thereby impeding the efficacy of {adaptation}; \textbf{(2) {Unsupervised nature.}} Throughout adaptation, ground-truth labels for input data are unavailable. Moreover, due to privacy concerns or legal restrictions, the source domain data may no longer be available and we only have access to the source pre-trained model; \textbf{(3) {Error accumulation and catastrophic forgetting.}} The dynamic distribution shift when adapting to continually changing target domains can render pseudo-labels unreliable and impede knowledge retention about the source domain, leading to error accumulation\cite{Wang_2022_CVPR,Wang_2024_WACV} and triggering catastrophic forgetting{~\cite{ijcai2023p183,liu2024vida,chen2022multi}}.


Addressing the first challenge, an intuitive solution is to employ a buffer to store high-value samples. However, existing methods that utilize such a buffer store source domain samples with ground-truth labels\cite{Dobler_2023_CVPR,Sojka_2023_ICCV}, which is difficult to satisfy in real-world scenarios and conflicts with the source-free constraint in the second challenge. Therefore, the issue of {effectively \emph{buffering} and {\emph{organizing}} high-value samples in an \emph{unsupervised} \emph{single-pass} data stream} has not been fully explored. {Current CTTA methods primarily address the third challenge, focusing mainly on} alignments in either the parameter space\cite{Brahma_2023_CVPR,niloy2024effective,press2024rdumb,Wang_2022_CVPR} or the feature space\cite{chakrabarty2023santa,Dobler_2023_CVPR} of the source model to mitigate knowledge forgetting. 

\begin{figure}[t]
    \centering
    \begin{subfigure}[b]{0.32\textwidth}
        \centering
        \includegraphics[height=2.5cm]{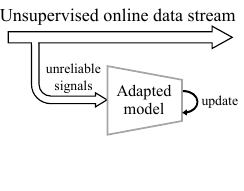}
        \caption{W/o a buffer}
        \label{fig:buffer_reshape1}
    \end{subfigure}
    \begin{subfigure}[b]{0.32\textwidth}
        \centering
        \includegraphics[height=2.5cm]{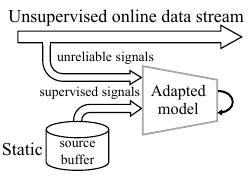}
        \caption{Static source buffer}
        \label{fig:buffer_reshape2}
    \end{subfigure}
    \begin{subfigure}[b]{0.32\textwidth}
        \centering
        \includegraphics[height=2.5cm]{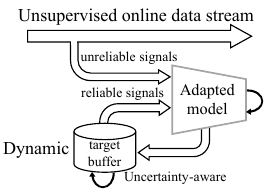}
        \caption{Our dynamic target buffer}
        \label{fig:buffer_reshape3}
    \end{subfigure}
    \caption{Comparison of different buffering mechanisms. (a) Without a buffer, only unreliable signals from the current unsupervised input data are available. (b) Some methods\cite{Dobler_2023_CVPR,Sojka_2023_ICCV} use a static source buffer that stores source samples with ground-truth labels to provide supervised signals. (c) We work on a dynamic target buffer, which can provide reliable signals for the adaptation process.}
    \label{fig:buffer_reshape}
\end{figure}

In this paper, we endeavor to address {these three} challenges simultaneously. To this end, we reshape the online data buffering and organizing mechanism for CTTA. {As shown in \cref{fig:buffer_reshape}}, unlike previous methods which directly sample from the source domain~\cite{Dobler_2023_CVPR,Sojka_2023_ICCV}, we are targeted at source-free, unsupervised, and online {buffering}. Through a specially designed {uncertainty-aware buffering approach}, we dynamically {evaluate and aggregate reliable information} in the {unsupervised}, {single-pass} data stream {(Challenges} \textbf{1} and \textbf{2}). 
{Based on this, we efficiently organize these samples to assist the model's unsupervised adaptation process. Utilizing pseudo-labels generated from these low-uncertainty samples, we consider them as anchors of classes and organize them into a class relation graph (CRG). To combat catastrophic forgetting under continually changing domains, we introduce a novel graph-based class relation preservation constraint to maintain the intrinsic class relations obtained on the source domain (Challenge \textbf{3}). 
Specifically, we penalize topological changes within the CRG to mitigate knowledge forgetting, while allowing reasonable vertex movements to enhance adaptability. Additionally, these samples are incorporated into a pseudo-target replay objective to reduce the risk of error accumulation from {misclassified} samples (Challenge \textbf{3}).} In summary, our main contributions are as follows:

\begin{itemize}

\item To address the three revealed challenges simultaneously, we reshape the online data buffering and organizing mechanism for CTTA. 

\item {To frame the buffering mechanism, we design an uncertainty-aware approach to {identify and aggregate reliable data} from unsupervised online data stream.}

\item {Through a well-designed graph-based class relation preservation constraint and a pseudo-target replay objective, we effectively mitigate catastrophic forgetting and error accumulation in CTTA.}

\item {We extensively evaluate our proposed method and the results on four challenging datasets show that the method clearly improves the CTTA performance on both segmentation and classification benchmarks, indicating its effectiveness and universality.}
\end{itemize}

\section{Related Work}
\subsection{Domain Adaptation}
Domain adaptation (DA)~\cite{zhang2023towards,patel2015visual} aims to generalize a model from a well-annotated source domain to an unlabeled target domain following the transductive learning principle. The underlying assumption in DA is that both labeled source domain and unlabeled target domain data are concurrently accessible. Consequently, the prevalent methods in DA encompass the utilization of domain adversarial training~\cite{cicek2019unsupervised,cui2020gradually,liu2019transferable,ganin2015unsupervised,tsai2018learning,hoffman2018cycada} or the minimization of domain discrepancy~\cite{long2015learning,long2017deep,peng2019moment}. These methods work towards harmonizing the feature distribution~\cite{ganin2015unsupervised,long2015learning,tsai2018learning} or the input space~\cite{hoffman2018cycada,yang2020fda} between the two domains.


\subsection{Continual Test-Time Adaptation}
\subsubsection{Test-Time Adaptation (TTA).}
TTA focuses on a more challenging setting after model deployment where only a pre-trained source model and unlabeled target data are available. Some approaches\cite{chen2022contrastive,liang2020we} focus on the offline setting, where multiple epochs of access to the entire set of test data are provided for adaptation, while others\cite{choi2022improving,boudiaf2022parameter,wang2023feature,liu2021ttt} focus on the online setting with a single-pass data stream. To name a few, TENT\cite{wang2021tent} adapts the model by using entropy minimization loss to update a few trainable parameters in batch normalization layers. This approach has also inspired subsequent works such as EATA~\cite{niu2022efficient} and RoTTA~\cite{yuan2023robust} to investigate the robustness of normalized layers. Due to the domain gap, TSD\cite{wang2023feature} promotes feature alignment and uniformity from the perspective of feature revision, while TIPI\cite{nguyen2023tipi} proposes invariance regularizers to find the transformation space that can simulate the domain shift.

\subsubsection{Continual Test-Time Adaptation (CTTA).}
In recent years, CTTA has received increasing research attention\cite{Wang_2022_CVPR,ijcai2023p183,gan2023decorate,song2023ecotta} due to its practicality. CoTTA\cite{Wang_2022_CVPR} stands out as the first method specifically tailored to address this task. It utilizes weight-averaged and augmentation-averaged predictions to reduce error accumulation, while stochastically restoring the weights of a small fraction of neurons to the source model during each iteration to mitigate catastrophic forgetting. 
Building upon this, PETAL\cite{Brahma_2023_CVPR} further introduces a data-driven parameter recovery technique to align with the source model in the parameter space, while RMT\cite{Dobler_2023_CVPR} and SANTA\cite{chakrabarty2023santa} seek alignment with the source model in the feature space by optimizing an additional contrastive learning objective. Additionally, DSS~\cite{Wang_2024_WACV} considers sample security to mitigate the risk of using incorrect information.
Some approaches\cite{gan2023decorate,song2023ecotta,liu2024vida} also prevent forgetting by freezing a majority or all of the model parameters, but they require optimization along with the model during training using the source domain data, thereby relaxing the unsupervised challenge and not being truly source-free. Under our proposal, we simultaneously address three challenges in CTTA.

\begin{figure}[tb]
  \centering
  \includegraphics[height=6.0cm]{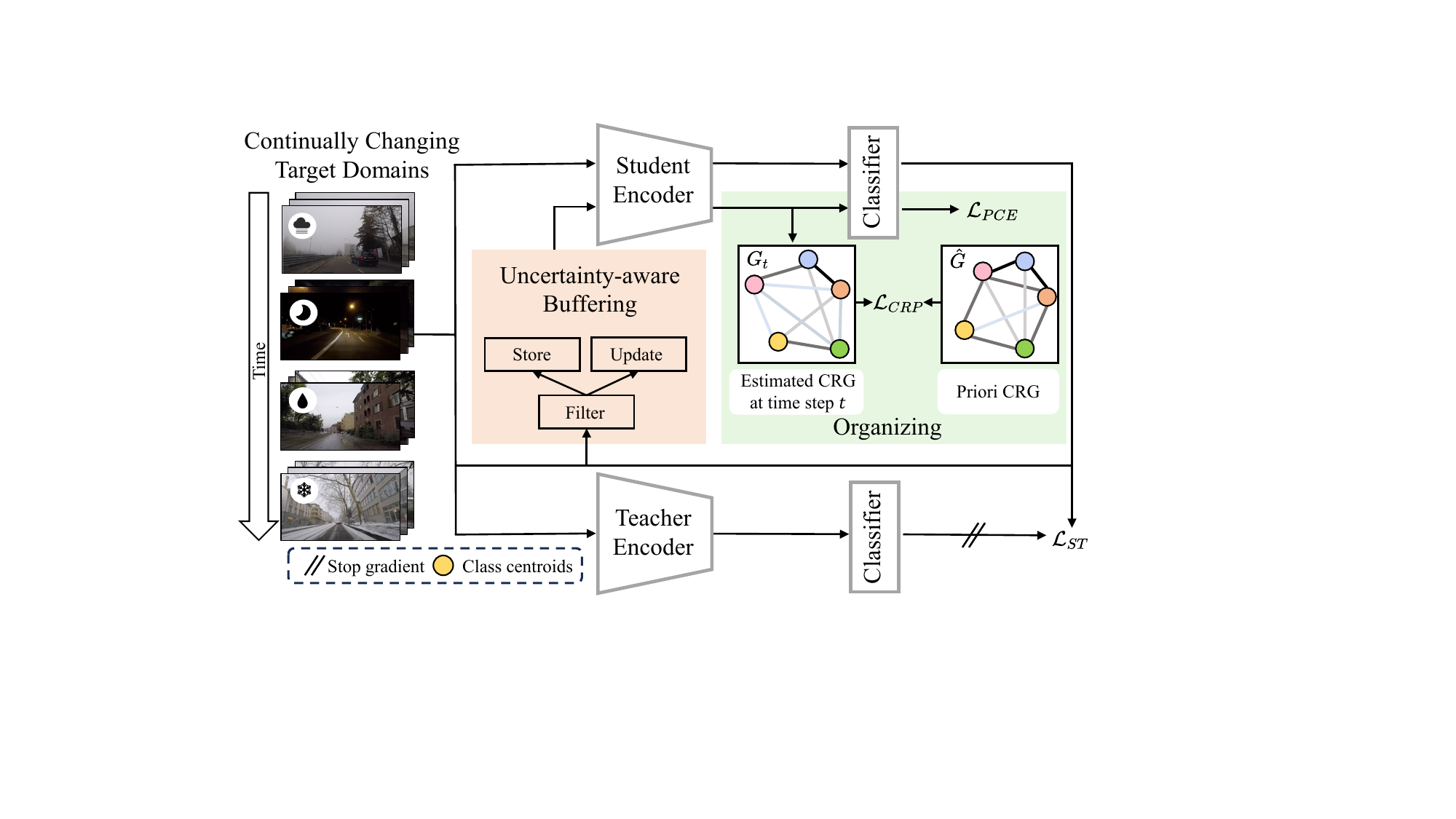}
  \caption{
    The framework of our proposed method. At time step $t$, the incoming data and sampled buffer data are used for adaptation. The proposed class relation preservation loss $\mathcal{L}_\textit{CRP}$ ensures topological consistency between the Class Relation Graph (CRG) $G_t$ estimated at the current time step $t$ and $\hat{G}$ from the source domain, thereby effectively preserving intrinsic semantic information. Concurrently, we integrate a pseudo-target replay loss $\mathcal{L}_\textit{PCE}$ to mitigate error accumulation.
    }
  \label{fig:method}
\end{figure}

\section{Methodology}
\subsection{Preliminaries and Overall Framework}
In continual test-time adaptation (CTTA), we are given a source model $f_{\theta_0}$ pre-trained from the labeled source domain $(\mathcal{X}^{S},\mathcal{Y}^{S})$ and a series of sequentially changing unlabeled target domains such as $\left [ \mathcal{X}^T_1, \mathcal{X}^T_2,\cdots, \mathcal{X}^T_n \right ] $. Each target domain shares the same label space as the source domain (for close-set adaptation). Note that the model has no information about when the domain changes happen during the adaptation process. The goal of CTTA is to adapt the model to a continually changing unsupervised online data stream of target domains without source data. 
To be more specific, at inference time step $t$, the unlabeled target test data {$x(t)$} is given to the model $f_{\theta_t}$, and the model needs to make predictions and adapt itself accordingly ($f_{\theta_t} \to f_{\theta_{t+1}}$) for the upcoming test inputs {$x(t+1)$}.

We adopt a teacher-student framework following previous works~\cite{Dobler_2023_CVPR,Brahma_2023_CVPR,Wang_2022_CVPR} as illustrated in \cref{fig:method}. Leveraging the proposed uncertainty-aware {buffering mechanism}, our method dynamically evaluates and buffers significant high-certainty samples in the unsupervised online data stream, providing reliable information for the model's adaptation process. Building upon this, we further introduce a novel class relation preservation constraint ($\mathcal{L}_\textit{CRP}$) for preserving domain-invariant semantic information in dynamic environments, effectively mitigating the catastrophic forgetting problem. Additionally, we address error accumulation through a pseudo-target replay objective ($\mathcal{L}_\textit{PCE}$) to minimize the impact of miscalibration. A self-training loss ($\mathcal{L}_\textit{ST}$) is used to stabilize the adaptation process of the student model (adapted model). In the adopted framework, both the teacher model $f_{\theta^{\prime}}$ and the student model $f_{\theta}$ are initialized with the source model $f_{\theta_0}$. Subsequently, the teacher model is continuously updated by the exponential moving average of the student model. We elaborate our method in the following subsections.



\subsection{Uncertainty-aware Buffering}\label{uncertainty-aware}
In the unsupervised online data stream within CTTA, the model necessitates adaptation utilizing the data at hand, typically characterized by a scarcity of information. While certain methods\cite{Dobler_2023_CVPR,Sojka_2023_ICCV} endeavor to incorporate additional \emph{supervised} information by formulating a static buffer from the source domain, fulfilling this requirement proves challenging within practical application scenarios. Conversely, our proposition involves sampling and buffing throughout the \emph{unsupervised} adaptation process by a designed uncertainty-aware approach.

At the outset, the dynamic buffer is initialized as empty. As the adaptation progresses, the buffer undergoes continual updates to reflect the evolving nature of the data stream encountered. Specifically, at any given inference time step $t$, the model calculates the uncertainty associated with the current batch of samples. The quantification of sample uncertainty manifests as the entropy of the sample, denoted as:
\begin{equation}
  H(x(t))=-f_{\theta_{t}}(x(t))^\top(\log{f_{\theta_{t}}(x(t))}),
  \label{eq:uncertainty}
\end{equation}
where $f_{\theta_{t}}(x(t))$ corresponds the softmax predictions of the student model for the current input data. Samples exhibiting an entropy value less than a predefined threshold $H_0$ are deemed to possess a high degree of certainty and are thus considered candidates for inclusion in the dynamic buffer. 
$H_0$ is defined as:
\begin{equation}
  H_0=\alpha \ln C,
  \label{eq:uncertainty_threshold}
\end{equation}
where $\alpha$ serves as a threshold coefficient and $C$ denotes the total number of classes in the dataset. The process of {storing} samples into the buffer, or updating existing entries, is contingent upon the buffer's capacity. High-certainty samples identified from the present batch, along with their derived pseudo-labels from the teacher model, are added to the buffer if space permits. Conversely, in scenarios where the buffer is at capacity, these new samples replace the ones exhibiting the highest uncertainty to maintain an optimal blend of timeliness and reliability. Upon completion of these updates, the resulting buffer $\mathcal{M}_{t}$ stands ready to facilitate the model's adaptation in subsequent iterations. 

Notably, our approach to buffering is entirely devoid of reliance on source data, underscoring its \emph{source-free} nature. 
This strategy enables the model to discern and prioritize data based on their uncertainty. By dynamically buffering high-certainty samples within the data stream, our approach not only circumvents the limitations posed by reliance on static source buffers but also enriches the adaptation process with valuable, reliable data. {The intuitive idea is to utilize CE loss to replay samples in the buffer, thereby emphasizing reliable data and effectively suppressing the risk and impact of miscalibration due to error accumulation.} 
In each test-time adaptation iteration, we randomly sample examples and labels $(\tilde{x}(t), \tilde{y}(t))$ from the buffer. The number of sampled examples is equal to the batch size and the corresponding pseudo-target replay loss is:
\begin{equation}
  \mathcal{L}_\textit{PCE} = -\sum_{c=1}^{C}\tilde{y}_{c}(t)\log{\tilde{p}_{c}},
  \label{eq:pce}
\end{equation}
where $\tilde{p}$ is the softmax prediction of the student model for $\tilde{x}(t)$. {For simplicity, we omit the summation of all samples in the batch.}
Moreover, these {sampled examples} also serve as anchors and are organized in a class relation graph. On this basis, we introduce a novel class relation preservation constraint in \cref{crp} to preserve domain-invariant knowledge and overcome forgetting.

\subsection{Class Relation Preservation Constraint}\label{crp}
In online scenarios where domains are constantly changing, neural networks tend to “take shortcuts” and focus on simplistic features~\cite{geirhos2020shortcut,Wei_2023_ICCV} from the single-pass data stream. This phenomenon results in the model's forgetting of intrinsic semantic information. In this work, we propose to overcome catastrophic forgetting by preserving the domain-invariant class relation, which is consistent across domains regardless of distribution discrepancy~\cite{zhang2023class}. Therefore, it is reasonable to utilize the intrinsic class relation obtained on the source domain with large labeled samples to constrain the model adaptation process on target domains.

{We describe here how to obtain the defined class relation graph (CRG)}. For computational stability, we normalize the feature space and adopt the cosine similarity metric. 
Let $ \bar{\cdot}$ denotes the normalization operation, where $ \bar{f} =f/\lVert f \rVert$. Given the normalized feature space $ \bar{\mathcal{F}}$, the CRG $G$ is defined as $G = \left \langle V, E \right \rangle $, where $V=\left \{ \bar{v}_1,\cdots , \bar{v}_C | \bar{v}_i \in \bar{\mathcal{F}} \right \}  $ is the set of $C$ vertices representative of $ \bar{\mathcal{F}}$, and $E$ is the edge set describing the neighborhood relations of vertices in $V$. Each edge $e_{ij}$ is assigned with a weight $s_{ij}$, which is the similarity between {$\bar{v}_i$ and $\bar{v}_j$: $s_{ij} = \bar{v}_i^\top\bar{v}_j$.}
Each vertex $\bar{v}_i$ is the centroid vector {of class $i$}.
The intrinsic CRG $\hat{G}$ is obtained by modeling the source-domain class relation. We demonstrate in experiments that the construction of our CRG $\hat{G}$ can be derived from different data conditions, including class prototypes and classifier weights.

At time step $t$, given CRG $G_t = \left \langle V_t, E_t \right \rangle $, when catastrophic forgetting occurs, $G_t$ is distorted as the semantic relations among categories are confused. To mitigate forgetting, the model should be adapted to strive towards preserving the \emph{topology} in $G_t$ the same as that in $\hat{G}$. This is achieved by restricting the neighboring relations of vertices described by the weights of edges (\ie, similarity between classes) in $E_t$. One way to do this is to keep the ranking of the edge weights constant during the adaptation process. However optimizing a non-smooth global ranking is difficult and inefficient~\cite{burges2006learning,tao2020topology}. An alternative and more feasible approach involves assessing the extent of change in neighboring relations{~\cite{tao2020bi,dong2021few}} by evaluating the correlation between the initial edge weights and their current observation. A lower correlation indicates a higher probability that the rank of an edge has changed during adaptation, and this should be penalized. On this basis, we define the Class Relation Preservation constraint (CRP) as follows:

\begin{equation}
  \mathcal{L}_\textit{CRP} = - \frac{\sum_{i,j}^{C} s_{ij} \tilde{s}_{ij} }{\sqrt{\sum_{i,j}^{C} s_{ij} ^2} \sqrt{\sum_{i,j} \tilde{s}_{ij} ^2}},
  \label{eq:class_relation_preservation}
\end{equation}
where $S=\left \{ s_{ij} | 1\le i,j\le C \right \} $ and $\tilde{S} =\left \{ \tilde{s}_{ij} | 1\le i,j\le C \right \} $ are the sets of the
initial and observation values of the edge weights in $\hat{E}$ and $E_t$, respectively. 

To achieve end-to-end optimization, the ideal situation is using source-domain images with \emph{ground-truth} labels to obtain an estimate of the current CRG $G_t$ under time step $t$. That is, extracting features from these images by $f_{\theta_t}$, which in turn yields the class prototypes and ultimately the class relation. However, it is often impractical to store a certain number of source-domain samples in the source-free CTTA setting. In this work, we leverage reliable data from our proposed uncertainty-aware buffering mechanism to accurately estimate CRG $G_t$ and achieve end-to-end optimization in an unsupervised test data stream.

From the perspective of viewing $S$ as a vector, the proposed CRP term constrains the consistency of the direction between two vectors during the adaptation process. Opting for the origin moment instead of the central moment allows the CRP term to prioritize direction consistency between vectors. This makes it more robust to scale variations or magnitude disparities in edge weights, which may arise from shifts in data distribution.
Some previous methods~\cite{Dobler_2023_CVPR,chakrabarty2023santa} directly penalize the movement of target domain samples in the feature space relative to the source prototypes. This can be broadly interpreted as penalizing the movement of corresponding elements between $\hat{V}$ and $V_t$ in our defined CRG. However, in addition to suffering from classification errors when assigning source prototypes, constraints on features can also impede the effectiveness of adaptation in target domains. Instead of penalizing the movement of vertices in G in the feature space, the proposed CRP term penalizes the change of topological relations between vertices, while allowing reasonable movement of vertices to enhance adaptation. To summarize, by preserving the domain-invariant class relation during the model adaptation process, the proposed CRP term effectively overcomes catastrophic forgetting caused by semantic confusion.

\subsection{Optimization Objective}
At time step $t$, the teacher model $f_{\theta_{t}^{\prime}}$ first generates pseudo-labels to help the learning process of the student model $f_{\theta_{t}}$. Specifically, we compute the symmetric cross-entropy loss~\cite{Dobler_2023_CVPR} between the outputs of the student model and the pseudo-labels. The self-training loss can be formulated as follows:
\begin{equation}
  \mathcal{L}_\textit{ST} = -\sum_{c=1}^{C} q_c\log{p_c} -\sum_{c=1}^{C} p_c\log{q_c},
  \label{eq:self_training}
\end{equation}
where $q$ is the softmax prediction of the teacher model, and $p$ is the softmax prediction of the student model, with $C$ being the total number of classes. Finally, the total loss function $\mathcal{L}_{T}$ can be formulated as follows:
\begin{equation}
  \mathcal{L}_{T} = \mathcal{L}_\textit{ST} + \mathcal{L}_\textit{PCE} + \lambda_{CRP} \mathcal{L}_\textit{CRP},
  \label{eq:total}
\end{equation}
where $\lambda_{CRP}$ is a trade-off hyperparameter.

\section{Experiments}
We conduct comprehensive experiments under the CTTA setting as in~\cite{Wang_2022_CVPR,Dobler_2023_CVPR}, encompassing {three mainstream datasets ImageNet-to-ImageNet-C, CIFAR10-to-CIFAR10C, and CIFAR100-to-CIFAR100C for image classification, as well as Cityscapes-to-ACDC for semantic segmentation.} 
\subsection{Datasets and Experimental Settings}
\subsubsection{CIFAR10C, CIFAR100C, ImageNet-C.} These datasets are originally developed to evaluate the robustness of classification networks~\cite{hendrycks2018benchmarking}. Each dataset contains 15 distinct types of corruption at varying severity levels ranging from 1 to 5. These corruptions are applied to the test and validation images of CIFAR and ImageNet, respectively. We employ the clean training sets on CIFAR10, CIFAR100, and ImageNet as source domain datasets, while CIFAR10C, CIFAR100C, and ImageNet-C serve as the corresponding target domain datasets. Following the continuous benchmark setup in~\cite{Wang_2022_CVPR}, we sequentially adapt the source model to 15 target domains under the largest severity level 5. The entire adaptation process is performed in an online manner, without assuming knowledge of when the domain changes. For each corruption, we utilize 10,000 images for datasets derived from CIFAR and 5,000 images for ImageNet-C.

\subsubsection{Cityscapes-to-ACDC.} This dataset is designed for the continual semantic segmentation task~\cite{Wang_2022_CVPR} in autonomous driving. The Cityscapes dataset~\cite{Cordts_2016_CVPR} is used as the source domain to provide an off-the-shelf pre-trained segmentation model, while the Adverse Conditions Dataset (ACDC)~\cite{sakaridis2021acdc}, which contains images collected under four different weather conditions: Fog, Night, Rain, and Snow, is used to represent the changing target domains. 400 unlabeled images from each condition are used for adaptation. Note that the ACDC dataset shares the same semantic classes as the evaluation classes of Cityscapes. To simulate continual distribution shifts encountered in real-world scenarios, we cyclically repeat the same sequence of target domains ten times (\eg, in total 40: Fog$\to$Night$\to$Rain$\to$Snow$\to$Fog$\to\dots$ ). This also provides a long-term perspective on the performance evaluation of different adaptation methods.

\subsubsection{Implementation Details.} To ensure consistency and comparability, we follow the implementation details from previous works~\cite{Dobler_2023_CVPR,Wang_2022_CVPR}. We adopt pre-trained WideResNet-28~\cite{zagoruyko2016wide} model for CIFAR10-to-CIFAR10C, pre-trained ResNeXt-29~\cite{xie2017aggregated} model for CIFAR100-to-CIFAR100C, and standard pre-trained ResNet-50 model for ImageNet-to-ImagenetC from Robustbench~\cite{croce2021robustbench} for all methods. Our batch size is set to 64 for ImageNet-C and 200 for other classification experiments. We evaluate the model using online predictions obtained immediately as the data is encountered. For segmentation CTTA, we employ a transformer-based architecture Segformer-B5~\cite{xie2021segformer} trained on Cityscapes as our segmentation model, and use down-sampled images from ACDC with a resolution of 960$\times$540 as network inputs. We utilize the Adam optimizer with a learning rate of 1e-4 and set the batch size to 1. 
To generate robust pseudo-labels, we use multi-scale input with flipping as the augmentation method following~\cite{Wang_2022_CVPR} for the segmentation experiments. When obtaining the intrinsic class relation $\hat{G}$, we utilize the weights of the classifier as proxies for classes, whereas in classification scenarios where feature prototypes are easily accessible\cite{Dobler_2023_CVPR,chakrabarty2023santa}, we utilize the prototypes to do so. All experiments are conducted using an NVIDIA RTX4090 GPU. 

\begin{table}[tb]
\belowrulesep=0pt
\aboverulesep=0pt
\normalsize
\caption{Semantic segmentation results (mIoU in \%) on the Cityscapes-to-ACDC CTTA task. The four test conditions are repeated ten times to evaluate the long-term adaptation performance. For brevity, we only show the continual adaptation results solely in the first, fourth, seventh, and final rounds. All results are evaluated based on the Segformer-B5 architecture. Bold text indicates the best performance.}
\label{tab:segmentation}
\centering
\resizebox{0.95\textwidth}{!}{
\begin{tabular}{l|ccccc|ccccc|ccccc|ccccc|c} 
\toprule
Time & \multicolumn{20}{c|}{$t\xrightarrow{\makebox[\dimexpr 40\width][c]{\quad}}$} & ~ \\
\midrule
{Round} & \multicolumn{5}{l|}{1} & \multicolumn{5}{l|}{4} & \multicolumn{5}{l|}{7} & \multicolumn{5}{l|}{10} & All  \\ 
\midrule
\multirow{3}*{Condition} & \multirow{3}*{\rotatebox{75}{Fog}} & \multirow{3}*{\rotatebox{75}{Night}} & \multirow{3}*{\rotatebox{75}{Rain}} & \multirow{3}*{\rotatebox{75}{Snow}} & \multirow{3}*{\rotatebox{75}{Mean}} & \multirow{3}*{\rotatebox{75}{Fog}} & \multirow{3}*{\rotatebox{75}{Night}} & \multirow{3}*{\rotatebox{75}{Rain}} & \multirow{3}*{\rotatebox{75}{Snow}} & \multirow{3}*{\rotatebox{75}{Mean}} & \multirow{3}*{\rotatebox{75}{Fog}} & \multirow{3}*{\rotatebox{75}{Night}} & \multirow{3}*{\rotatebox{75}{Rain}} & \multirow{3}*{\rotatebox{75}{Snow}} &\multirow{3}*{\rotatebox{75}{Mean}} & \multirow{3}*{\rotatebox{75}{Fog}} & \multirow{3}*{\rotatebox{75}{Night}} & \multirow{3}*{\rotatebox{75}{Rain}} & \multirow{3}*{\rotatebox{75}{Snow}} & \multirow{3}*{\rotatebox{75}{Mean}} & \multirow{3}*{Mean$\uparrow$} \\
~ & ~ & ~ & ~ & ~ & ~ & ~ & ~ & ~ & ~ & ~ & ~ & ~ & ~ & ~ & ~ & ~ & ~ & ~ & ~ & ~ & ~ \\
~ & ~ & ~ & ~ & ~ & ~ & ~ & ~ & ~ & ~ & ~ & ~ & ~ & ~ & ~ & ~ & ~ & ~ & ~ & ~ & ~ & ~ \\
\midrule
Source & 69.1 & 40.3 & 59.7 & 57.8 & 56.7 & 69.1 & 40.3 & 59.7 & 57.8 & 56.7 & 69.1 & 40.3 & 59.7 & 57.8 & 56.7 & 69.1 & 40.3 & 59.7 & 57.8 & 56.7 & 56.7 \\
BN Stats Adapt & 62.3 & 38.0 & 54.6 & 53.0 & 52.0 & 62.3 & 38.0 & 54.6 & 53.0 & 52.0 & 62.3 & 38.0 & 54.6 & 53.0 & 52.0 & 62.3 & 38.0 & 54.6 & 53.0 & 52.0 & 52.0 \\
TENT-cont.\cite{wang2021tent} & 69.0 & 40.2 & 60.1 & 57.3 & 56.7 & 66.5 & 36.3 & 58.7 & 54.0 & 53.9 & 64.2 & 32.8 & 55.3 & 50.9 & 50.8 & 61.8 & 29.8 & 51.9 & 47.8 & 47.8 & 52.3 \\
CoTTA\cite{Wang_2022_CVPR} & 70.9 & 41.2 & 62.4 & 59.7 & 58.6 & 70.9 & 41.2 & 62.4 & 59.7 & 58.6 & 70.9 & 41.2 & 62.4 & 59.7 & 58.6 & 70.9 & 41.2 & 62.4 & 59.7 & 58.6 & 58.6 \\
Ours & 71.2 & 42.3 & 65.0 & 62.0 & \textbf{60.1} & 72.8 & 43.6 & 66.7 & 63.3 & \textbf{61.6} & 72.5 & 42.5 & 66.8 & 63.3 & \textbf{61.3} & 72.5 & 42.9 & 66.7 & 63.0 & \textbf{61.3} & \textbf{61.3} \\
\bottomrule
\end{tabular}
}
\end{table}

\subsection{Experiments on Segmentation CTTA}
We report results based on the mean intersection over union (mIoU) metric for the complex continual test-time semantic segmentation Cityscapes-to-ACDC task. As detailed in \cref{tab:segmentation}, a discernible decline in the performance of TENT across long-term sequences is observed, highlighting the pervasive issue of model degradation over time. Conversely, our proposed method effectively mitigates catastrophic forgetting and error accumulation, thereby consistently maintaining a high mean mIoU level across repeated sequences of target domains.
Particularly noteworthy is our method's achievement of the highest mIoU score, showcasing an impressive relative improvement of 4.6\% in mIoU compared to the previous sota method CoTTA~\cite{Wang_2022_CVPR}, averaged across ten rounds. This significant improvement demonstrates the ability of our method to adapt to dynamic environments within dense prediction tasks over the long term. Additionally, there is a trend of incremental improvement in the average mIoU metric during the initial rounds, indicating the capacity of our method to extract and leverage the reliable information embedded within the unsupervised data stream. This capability facilitates further refinement of the adaptation performance.
Comprehensive results are available in the supplementary material.

\begin{figure}[tb]
  \centering
  \includegraphics[height=5.35cm]{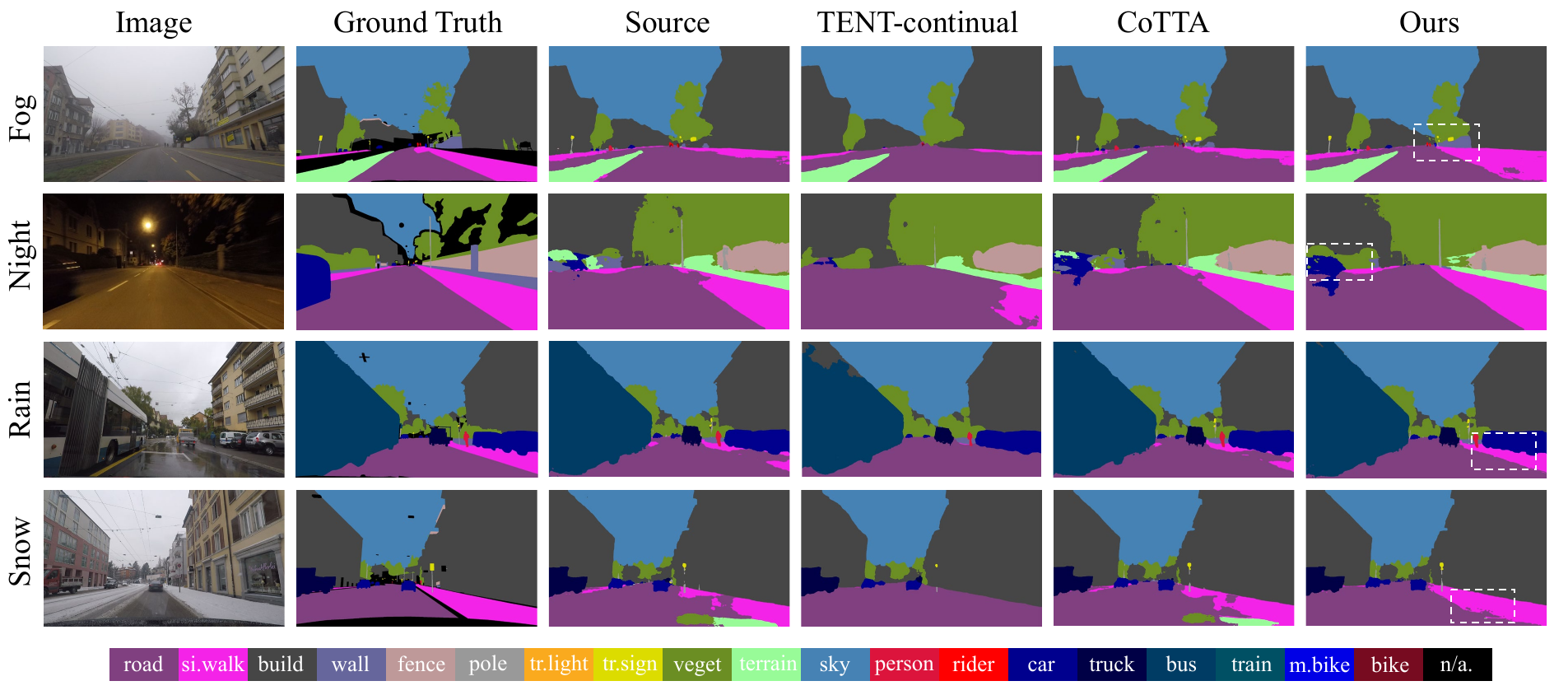}
  \caption{
    Qualitative comparison of semantic segmentation on the Cityscapes-to-ACDC CTTA task. The results for all methods come from the final round.
  }
  \label{fig:seg}
\end{figure}

For a more intuitive validation of the effectiveness of our method, we conduct qualitative comparison experiments and 
show the results obtained through the CTTA process in \cref{fig:seg}. Compared to other state-of-the-art methods, our results exhibit superior clarity and accuracy, yielding the best segmentation maps in all four target domains. Our method provides consistent improvements for most of the categories (as shown by the white boxes), while the comparison methods mostly suffer from severe semantic confusion. These observations demonstrate the effectiveness of our method in complex real-world environments.

\subsection{Experiments on Classification CTTA}
\begin{table}[ht]
\belowrulesep=0pt
\aboverulesep=0pt
\normalsize
\caption{Classification error rate (\%, lower is better) for the standard ImageNet-to-ImageNet-C CTTA task. All results are evaluated with the largest corruption severity level 5 in an online manner. Bold text indicates the best performance.}%
\label{tab:ImageNet-C}
\centering
\resizebox{0.95\textwidth}{!}{
\begin{tabular}{l|ccccccccccccccc|c} 
\toprule
Time & \multicolumn{15}{c|}{$t\xrightarrow{\makebox[\dimexpr 29\width][c]{\quad}}$} & ~ \\
\midrule
\multirow{4}{*}{Method} & \multirow{4}*{\rotatebox{75}{Gaussian}} & \multirow{4}*{\rotatebox{75}{shot}} & \multirow{4}*{\rotatebox{75}{impulse}} & \multirow{4}*{\rotatebox{75}{defocus}} & \multirow{4}*{\rotatebox{75}{glass}} & \multirow{4}*{\rotatebox{75}{motion}} & \multirow{4}*{\rotatebox{75}{zoom}} & \multirow{4}*{\rotatebox{75}{snow}} & \multirow{4}*{\rotatebox{75}{frost}} & \multirow{4}*{\rotatebox{75}{fog}} & \multirow{4}*{\rotatebox{75}{brightness}} & \multirow{4}*{\rotatebox{75}{contrast}} & \multirow{4}*{\rotatebox{75}{elastic}} & \multirow{4}*{\rotatebox{75}{pixelate}} & \multirow{4}*{\rotatebox{75}{jpeg}} & \multirow{4}*{{Mean$\downarrow$}} \\
~ & ~ & ~ & ~ & ~ & ~ & ~ & ~ & ~ & ~ & ~ & ~ & ~ & ~ & ~ & ~ & ~ \\
~ & ~ & ~ & ~ & ~ & ~ & ~ & ~ & ~ & ~ & ~ & ~ & ~ & ~ & ~ & ~ & ~ \\
~ & ~ & ~ & ~ & ~ & ~ & ~ & ~ & ~ & ~ & ~ & ~ & ~ & ~ & ~ & ~ & ~ \\
\midrule
Source & 97.8 & 97.1 & 98.2 & 81.7 & 89.8 & 85.2 & 78.0 & 83.5 & 77.1 & 75.9 & 41.3 & 94.5 & 82.5 & 79.3 & 68.6 & 82.0 \\
BN Stats Adapt & 85.0 & 83.7 & 85.0 & 84.7 & 84.3 & 73.7 & 61.2 & 66.0 & 68.2 & 52.1 & 34.9 & 82.7 & 55.9 & 51.3 & 59.8 & 68.6 \\
TENT-cont.~\cite{wang2021tent} & 81.6 & 74.6 & 72.7 & 77.6 & 73.8 & 65.5 & 55.3 & 61.6 & 63.0 & 51.7 & 38.2 & 72.1 & 50.8 & 47.4 & 53.3 & 62.6 \\
CoTTA\cite{Wang_2022_CVPR} & 84.7 & 82.1 & 80.6 & 81.3 & 79.0 & 68.6 & 57.5 & 60.3 & 60.5 & 48.3 & 36.6 & 66.1 & 47.2 & 41.2 & 46.0 & 62.7 \\
RMT\cite{Dobler_2023_CVPR} & 80.2 & 76.4 & 74.5 & 77.1 & 74.4 & 66.2 & 57.6 & 57.0 & 59.1 & 48.0 & 39.1 & 60.6 & 47.3 & 42.5 & 43.4 & 60.2 \\
DSS\cite{Wang_2024_WACV} & 84.6 & 80.4 & 78.7 & 83.9 & 79.8 & 74.9 & 62.9 & 62.8 & 62.9 & 49.7 & 37.4 & 71.0 & 49.5 & 42.9 & 48.2 & 64.6 \\
Ours & 78.5 & 75.3 & 73.0 & 75.7 & 73.1 & 64.5 & 56.0 & 55.8 & 58.1 & 47.6 & 38.5 & 58.5 & 46.1 & 42.0 & 43.4 & \textbf{59.0} \\
\bottomrule
\end{tabular}
}
\end{table}

\subsubsection{ImageNet-to-ImageNet-C.} Given the source model pre-trained on ImageNet, we sequentially execute CTTA on the 15 domains of ImageNet-C. As shown in \cref{tab:ImageNet-C}, when directly testing the source model on target domains, the average classification error rate is as high as 82.0\%, which demonstrates the need for adaptation. Leveraging the batch normalization statistics from the input data of the current iteration yielded the BN Stats Adapt method~\cite{li2016revisiting,schneider2020improving}, a relatively straightforward approach that reduces the average classification error rate to 68.6\%. While the TENT-based method~\cite{wang2021tent} initially helps the model to continuously adapt to target domains, it is prone to severe catastrophic forgetting and error accumulation over long periods.
CoTTA~\cite{Wang_2022_CVPR} and RMT~\cite{Dobler_2023_CVPR} align to the source domain either in parameter space or feature space, respectively, intending to mitigate the problem of catastrophic forgetting, and both have made some progress. DSS~\cite{Wang_2024_WACV} has noted the existence of high-quality and low-quality samples in the data stream, but it lacks further exploration on how to organize and utilize this information.

In our method, we reshape the online data buffering mechanism in CTTA, enabling dynamic aggregation of high-value samples from the unsupervised data stream. Based on this, we introduce a novel graph-based class relation preservation constraint alongside a pseudo-target replay objective.
As a result, our method achieves a substantial reduction in the average classification error rate from 64.6\% observed in the DSS method to 59.0\%.
Notably, our method outperforms all previous methods and reduces the error rate by a relative 2.0\% compared to the previous sota method, averaged across 15 domains. Moreover, our method demonstrates the best performance on the vast majority of corruption types, highlighting its effectiveness in mitigating catastrophic forgetting and error accumulation.

\subsubsection{CIFAR10-to-CIFAR10C, CIFAR100-to-CIFAR100C.}
\begin{table}[tb]
\belowrulesep=0pt
\aboverulesep=0pt
\normalsize
\caption{Classification error rate (\%, lower is better) for the standard CTTA tasks on
CIFAR10-to-CIFAR10C, CIFAR100-to-CIFAR100C. All results are evaluated with the largest corruption severity level 5 in an online manner. {Bold} text indicates the best.}
\label{tab:cifar}
\centering
\resizebox{0.95\textwidth}{!}{
\begin{tabular}{l|l|ccccccccccccccc|c} 
\toprule
~ & Time & \multicolumn{15}{c|}{$t\xrightarrow{\makebox[\dimexpr 29\width][c]{\quad}}$} & ~ \\
\midrule
~ & \multirow{4}{*}{Method} & \multirow{4}*{\rotatebox{75}{Gaussian}} & \multirow{4}*{\rotatebox{75}{shot}} & \multirow{4}*{\rotatebox{75}{impulse}} & \multirow{4}*{\rotatebox{75}{defocus}} & \multirow{4}*{\rotatebox{75}{glass}} & \multirow{4}*{\rotatebox{75}{motion}} & \multirow{4}*{\rotatebox{75}{zoom}} & \multirow{4}*{\rotatebox{75}{snow}} & \multirow{4}*{\rotatebox{75}{frost}} & \multirow{4}*{\rotatebox{75}{fog}} & \multirow{4}*{\rotatebox{75}{brightness}} & \multirow{4}*{\rotatebox{75}{contrast}} & \multirow{4}*{\rotatebox{75}{elastic}} & \multirow{4}*{\rotatebox{75}{pixelate}} & \multirow{4}*{\rotatebox{75}{jpeg}} & \multirow{4}*{{Mean$\downarrow$}} \\
~ & ~ & ~ & ~ & ~ & ~ & ~ & ~ & ~ & ~ & ~ & ~ & ~ & ~ & ~ & ~ & ~ & ~ \\
~ & ~ & ~ & ~ & ~ & ~ & ~ & ~ & ~ & ~ & ~ & ~ & ~ & ~ & ~ & ~ & ~ & ~ \\
~ & ~ & ~ & ~ & ~ & ~ & ~ & ~ & ~ & ~ & ~ & ~ & ~ & ~ & ~ & ~ & ~ & ~ \\
\midrule
\multirow{9}*{\rotatebox{90}{CIFAR10C}} & Source & 72.3 & 65.7 & 72.9 & 46.9 & 54.3 & 34.8 & 42.0 & 25.1 & 41.3 & 26.0 & 9.3 & 46.7 & 26.6 & 58.5 & 30.3 & 43.5 \\
~ & BN Stats Adapt & 28.1 & 26.1 & 32.5 & 13.2 & 35.3 & 14.2 & 17.3 & 12.7 & 17.3 & 15.3 & 8.4 & 12.6 & 23.8 & 19.7 & 27.3 & 20.4 \\
~ & TENT-cont.~\cite{wang2021tent} & 24.8 & 20.6 & 28.6 & 14.4 & 31.1 & 16.5 & 14.1 & 19.1 & 18.6 & 18.6 & 12.2 & 20.3 & 25.7 & 20.8 & 24.9 & 20.7 \\
~ & CoTTA\cite{Wang_2022_CVPR} & 24.3 & 21.3 & 26.6 & 11.6 & 27.6 & 12.2 & 10.3 & 14.8 & 14.1 & 12.4 & 7.5 & 10.6 & 18.3 & 13.4 & 17.3 & 16.2 \\
~ & RMT\cite{Dobler_2023_CVPR} & 24.0 & 20.4 & 25.6 & 12.6 & 25.4 & 14.2 & 12.2 & 15.4 & 15.1 & 14.1 & 10.3 & 13.7 & 17.1 & 13.5 & 16.0 & 16.7 \\
~ & PETAL\cite{Brahma_2023_CVPR} & 23.4 & 21.1 & 25.7 & 11.7 & 27.2 & 12.2 & 10.3 & 14.8 & 13.7 & 12.7 & 7.4 & 10.5 & 18.1 & 13.4 & 16.8 & 16.0 \\
~ & DSS\cite{Wang_2024_WACV} & 24.1 & 21.3 & 25.4 & 11.7 & 26.9 & 12.2 & 10.5 & 14.5 & 14.1 & 12.5 & 7.8 & 10.8 & 18.0 & 13.1 & 17.3 & 16.0 \\
~ & SANTA\cite{chakrabarty2023santa} & 23.9 & 20.1 & 28.0 & 11.6 & 27.4 & 12.6 & 10.2 & 14.1 & 13.2 & 12.2 & 7.4 & 10.3 & 19.1 & 13.3 & 18.5 & 16.1 \\
~ & Ours & 23.6 & 19.9 & 26.0 & 11.8 & 25.3 & 13.2 & 10.9 & 14.3 & 13.5 & 12.7 & 9.0 & 11.9 & 17.1 & 12.7 & 15.9 & \textbf{15.8} \\

\midrule
\multirow{9}*{\rotatebox{90}{CIFAR100C}} & Source & 73.0 & 68.0 & 39.4 & 29.3 & 54.1 & 30.8 & 28.8 & 39.5 & 45.8 & 50.3 & 29.5 & 55.1 & 37.2 & 74.7 & 41.2 & 46.4 \\
~ & BN Stats Adapt & 42.1 & 40.7 & 42.7 & 27.6 & 41.9 & 29.7 & 27.9 & 34.9 & 35.0 & 41.5 & 26.5 & 30.3 & 35.7 & 32.9 & 41.2 & 35.4 \\
~ & TENT-cont.~\cite{wang2021tent} & 37.2 & 35.8 & 41.7 & 37.9 & 51.2 & 48.3 & 48.5 & 58.4 & 63.7 & 71.1 & 70.4 & 82.3 & 88.0 & 88.5 & 90.4 & 60.9 \\
~ & CoTTA\cite{Wang_2022_CVPR} & 40.1 & 37.7 & 39.7 & 26.9 & 38.0 & 27.9 & 26.4 & 32.8 & 31.8 & 40.3 & 24.7 & 26.9 & 32.5 & 28.3 & 33.5 & 32.5 \\
~ & RMT\cite{Dobler_2023_CVPR} & 40.5 & 36.1 & 36.3 & 27.7 & 33.9 & 28.5 & 26.4 & 29.0 & 29.0 & 32.5 & 25.1 & 27.4 & 28.2 & 26.3 & 29.3 & 30.4 \\
~ & PETAL\cite{Brahma_2023_CVPR} & 38.3 & 36.4 & 38.6 & 25.9 & 36.7 & {27.2} & 25.4 & 32.0 & 30.8 & 38.7 & 24.4 & 26.4 & 31.5 & 26.9 & 32.5 & 31.5 \\
~ & DSS\cite{Wang_2024_WACV} & 39.7 & 36.0 & 37.2 & 26.3 & 35.6 & 27.5 & 25.1 & 31.4 & 30.0 & 37.8 & 24.2 & 26.0 & 30.0 & 26.3 & 31.1 & 30.9 \\
~ & SANTA\cite{chakrabarty2023santa} & {36.5} & {33.1} & {35.1} & {25.9} & 34.9 & 27.7 & 25.4 & 29.5 & 29.9 & 33.1 & {23.6} & 26.7 & 31.9 & 27.5 & 35.2 & 30.4 \\
~ & Ours & 38.8 & 35.0 & 35.4 & 26.7 & {33.2} & 27.4 & {25.0} & {27.4} & {26.8} & {29.8} & 24.1 & {25.1} & {26.9} & {24.9} & {28.0} & \textbf{29.0} \\

\bottomrule
\end{tabular}
}
\end{table}

To further demonstrate the effectiveness of the proposed method, we evaluate it on CIFAR10-to-CIFAR10C and CIFAR100-to-CIFAR100C CTTA tasks that have different numbers of categories. 
The experimental results are summarized in \cref{tab:cifar}. Once again, our method outperforms all previous competing methods and reduces the average error rate to 15.8\% and 29.0\%, respectively. Compared with both the previous sota parameter space recovery methods (PETAL\cite{Brahma_2023_CVPR}) and feature space alignment methods (SANTA\cite{chakrabarty2023santa}), our method shows substantial improvements, demonstrating the superiority of the proposed class relation preservation mechanism in overcoming catastrophic forgetting. The consistency enhancement across datasets with varying levels of complexity proves the effectiveness and universality of our method.

\subsubsection{Experiments on Gradually Changing Setup.}
\begin{table}[tb]
\belowrulesep=0pt
\aboverulesep=0pt

\caption{Gradually changing CIFAR10-to-CIFAR10C results. The severity level changes gradually between the lowest and the highest.}
\label{tab:cifar10c_gradual}
\centering
\resizebox{0.95\textwidth}{!}{
\begin{tabular}{c|c|c|c|c|c|c|c} 
\toprule
Avg. Error (\%) & Source & BN Stats Adapt & TENT-cont.~\cite{wang2021tent} & CoTTA\cite{Wang_2022_CVPR} & RMT\cite{Dobler_2023_CVPR} & SANTA\cite{chakrabarty2023santa} & Our \\
\midrule
CIFAR10C & 24.7 & 13.7 & 20.4 & 10.9 & 10.4 & 10.7 & \textbf{9.9} \\
\bottomrule
\end{tabular}
}
\end{table}
Following~\cite{chakrabarty2023santa,Dobler_2023_CVPR,Wang_2022_CVPR}, we also experiment with a gradually changing setup. Different from the previous standard setup where the corruption type changes abruptly at the largest severity level, the change process under the gradually changing setup can be represented as follows:
$\underbrace{\ldots \to 2\to 1}_{\text{before}}\xrightarrow[type]{change}\underbrace{1\to 2\to 3\to 4\to 5\to 4\to 3\to 2\to 1}_{\text{current type, gradually changing severity}} \xrightarrow[type]{change}\underbrace{1\to 2\to \ldots}_{\text{after}}$, where each number denotes the severity of corruption. The experimental results are shown in \cref{tab:cifar10c_gradual}, and our proposed method also achieves superior performance in this setup, reducing the average error rate to 9.9\% on the gradual CIFAR10-to-CIFAR10C benchmark.


\subsection{Ablation Study}
\subsubsection{Effectiveness of Each Component.}
We conduct the ablation study on the CIFAR100-to-CIFAR100C scenario and evaluate the effectiveness of each component in \cref{tab:ab}. $Ex1$ denotes the performance using only self-training loss in the teacher-student framework, yielding a relatively high average error rate. From $Ex2$ to $Ex4$, we demonstrate the importance of our proposed uncertainty-aware buffering mechanism. In contrast, the reservoir and FIFO buffering mechanisms do not have the ability to actively perform information aggregation, and thus they provide only a negligible improvement when combined with our pseudo-target replay loss. $Ex5$ demonstrates the effectiveness of our proposed class relation preservation constraint. $Ex6$
shows a significant overall improvement, suggesting that our proposed components are complementary. 

\makeatletter\def\@captype{figure}\makeatother
\begin{minipage}{.54\textwidth} 
  \centering
  \includegraphics[height=2.6cm]{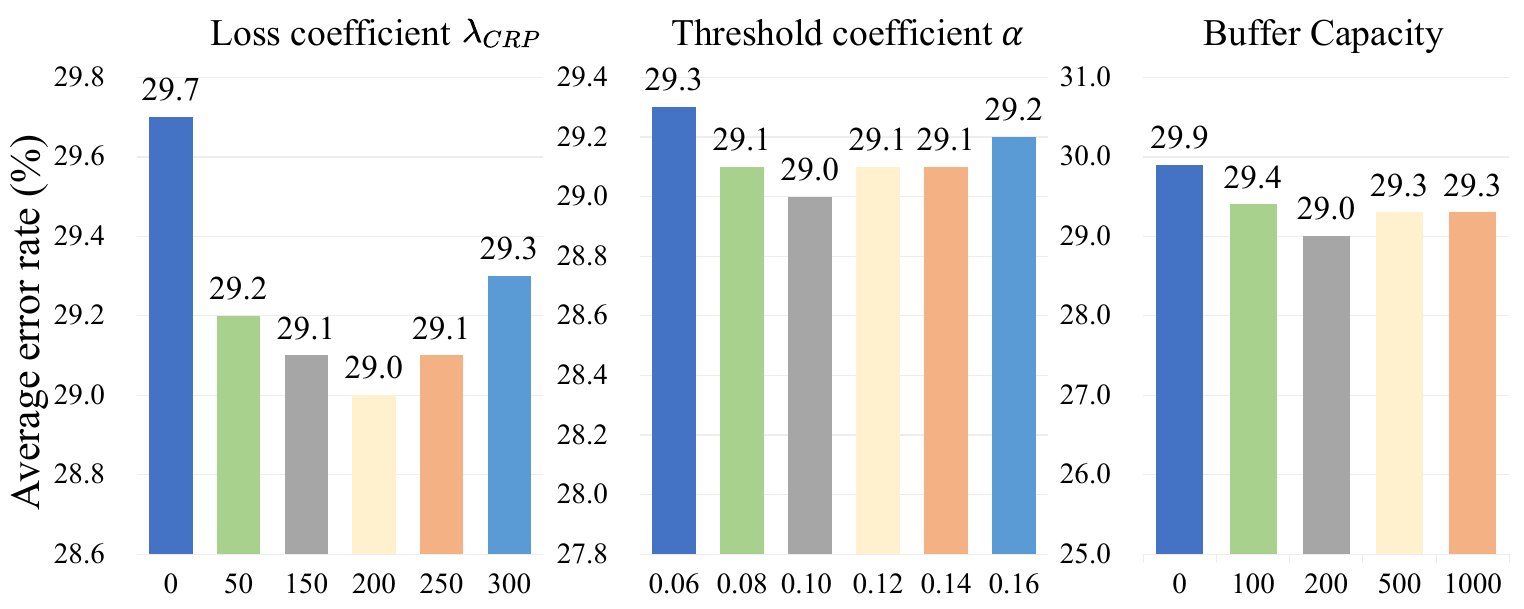}
  \caption{Hyperparameter sensitivity analysis.
  }
  \label{fig:subfigures}
\end{minipage}
\makeatletter\def\@captype{table}\makeatother
\begin{minipage}{.42\textwidth}
\belowrulesep=0pt
\aboverulesep=0pt
\caption{Ablation experiments.}
\label{tab:ab}
\centering
\resizebox{0.9\textwidth}{!}{
\begin{tabular}{c|ccc|c} 
\toprule
~ & Buffering & $\mathcal{L}_\textit{PCE}$ & $\mathcal{L}_\textit{CRP}$ & Mean$\downarrow$ \\
\midrule
$Ex_1$ & - & - & - & 30.7 \\
$Ex_2$ & Reservoir & \checkmark & - & 30.5 \\
$Ex_3$ & FIFO & \checkmark & - & 30.6 \\
$Ex_4$ & Uncertainty & \checkmark & - & 29.7 \\
$Ex_5$ & - & - & \checkmark & 30.0 \\
$Ex_6$ & Uncertainty & \checkmark & \checkmark & 29.0 \\
\bottomrule
\end{tabular}
}
\end{minipage}

\subsubsection{Hyperparameter Sensitivity.}
Here we analyze the sensitivity of hyperparameters within the CIFAR100-to-CIFAR100C scenario.
We consider three hyperparameters as illustrated in \cref{fig:subfigures}.
\textbf{Loss coefficient $\lambda_{CRP}$.} The hyperparameter $\lambda_{CRP}$ in \cref{eq:total} controls the strength of the proposed CRP constraint. We vary it within the range of $\left \{ 0, 50, 150, 200, 250, 300 \right \} $. It can be observed that increasing $\lambda_{CRP}$ over a reasonably broad range leads to a reduction in the average classification error rate of CTTA across 15 domains, indicating the effectiveness of CRP.
While excessively large values of $\lambda_{CRP}$ (\eg, 300) may compromise the model's adaptability. \textbf{Threshold coefficient $\alpha$.} The hyperparameter $\alpha$ in \cref{eq:uncertainty_threshold} is utilized to adjust the threshold for screening low-uncertainty samples in the streaming data. It is evident that it is stable over specific ranges and achieves the lowest average error rate among all domains at a value of 0.1. \textbf{Buffer capacity.} Here we examine the capacity of the buffer, which is employed to dynamically aggregate high-value samples from the online unsupervised data stream. As demonstrated in \cref{fig:subfigures}, our method achieves optimal results with a buffer capacity of 200, imposing only a negligible storage overhead.

\section{Conclusion}
In this paper, we {revisit} the key challenges in CTTA and address all of them simultaneously. To this end, we reshape the online data buffering and organizing mechanism, endowing our dynamic target buffer with the ability to aggregate reliable information in the unsupervised data stream.
Through our designed class relation preservation constraint and a pseudo-target replay objective, the problem of catastrophic forgetting and error accumulation is significantly alleviated. 
Extensive experiments demonstrate the effectiveness of the proposed method. 
{Regarding the adoption of uncertainty, our method employs the mainstream practice of sample entropy. In our framework, there is potential for investigating more diverse forms of uncertainty modeling, such as those based on energy functions. We leave this exploration to future work.}

\bibliographystyle{splncs04}
\bibliography{main}

\clearpage
\appendix

\setcounter{table}{6} 
\setcounter{figure}{4} 

In the supplementary material, we provide more analysis and experiments on this paper. In \cref{adaptarion_settings}, we provide a detailed explanation of different adaptation settings involved in \cref{tab:settings}. \cref{cam} provides additional class activation mapping (CAM) visualization results as supplementary empirical observations. \cref{seg_ctta} provides more details and comprehensive results of the semantic segmentation CTTA task. In \cref{ptta}, we provide an additional comparison with a recent approach that uses a buffer to solve temporal correlation, demonstrating the effectiveness of the buffering and organizing mechanism in our method in simultaneously addressing three key challenges in CTTA.
  
\section{Detailed Explanation on Different Adaptation Settings}\label{adaptarion_settings}
In \cref{tab:settings}, we compare different adaptation settings and distinguish them based on four aspects: the available data form of the source domain and target domain, the distribution and access mode of the target domain data stream during adaptation. A \emph{stationary} distribution means that the data on the target domain obeys the same distribution, while a \emph{continually changing} distribution indicates that the distribution of data on the target domain changes over time. Regarding the access mode, \emph{offline} means that the model has multiple epochs of access to the entire set of test data, while \emph{online} means that the input is a single-pass data stream and the model can only process the current batch of data at a time. In the following, we provide further explanation for each setting.
\paragraph{Fine-tuning (FT):} FT\cite{guo2019spottune,tajbakhsh2016convolutional} is a popular paradigm in transfer learning, where parameters of a source model are fine-tuned based on labeled data from the target domain. The data distribution on the target domain is stationary, and the model can have multiple rounds of access to the entire labeled target data.
\paragraph{Domain Generalization (DG):} DG\cite{lee2022cross,zhang2022mvdg,li2018domain} aims to train more generalizable neural networks from the source domain to improve the model's performance on unseen domains. Therefore, DG assumes a large number of labeled source domain samples for training while making no assumptions about the target domain.
\paragraph{Domain Adaptation (DA):} DA~\cite{hwang2022combating,lin2022prototype,li2021transferable} aims to generalize a source model from a well-annotated source domain to an unlabeled target domain. It jointly optimizes labeled source data and unlabeled target data to alleviate the distribution shift, where the adaptation is offline and requires multiple training epochs. 
\paragraph{Test-Time Training (TTT):} TTT\cite{liu2021ttt,sun2020test} focuses on online adaptation settings in the target domain, leveraging auxiliary self-supervised branches to update the model with current test data. However, it requires retraining the source model with source data to learn the self-supervised auxiliary branches, thus not directly compatible with existing pre-trained models.
\paragraph{Test-Time Adaptation (TTA):} TTA emphasizes settings after model deployment, where only a source model and unlabeled target data are available. Some methods focus on offline settings\cite{chen2022contrastive,liang2020we}, while others focus on online settings\cite{choi2022improving,boudiaf2022parameter,wang2023feature,liu2021ttt} with a single-pass data stream. All these approaches focus only on scenarios where the target domain is stationary, \ie, the target domain consists of only a single domain.
\paragraph{Continual Test-Time Adaptation (CTTA):} Given the rarity of encountering a single-domain shift in real-world scenarios, CTTA\cite{Wang_2022_CVPR,Brahma_2023_CVPR,Dobler_2023_CVPR,chakrabarty2023santa} has gained increasing attention due to its practicality. In this setting, the distribution of target domains is continually changing, aiming to evaluate the ability of methods to adapt to changing domains. As we reveal in the paper, the key challenges of this task are the online environment, the unsupervised nature, and the risk of error accumulation and catastrophic forgetting under continual domain shifts.

\begin{figure}[tb]
  \centering
  \includegraphics[height=6.0cm]{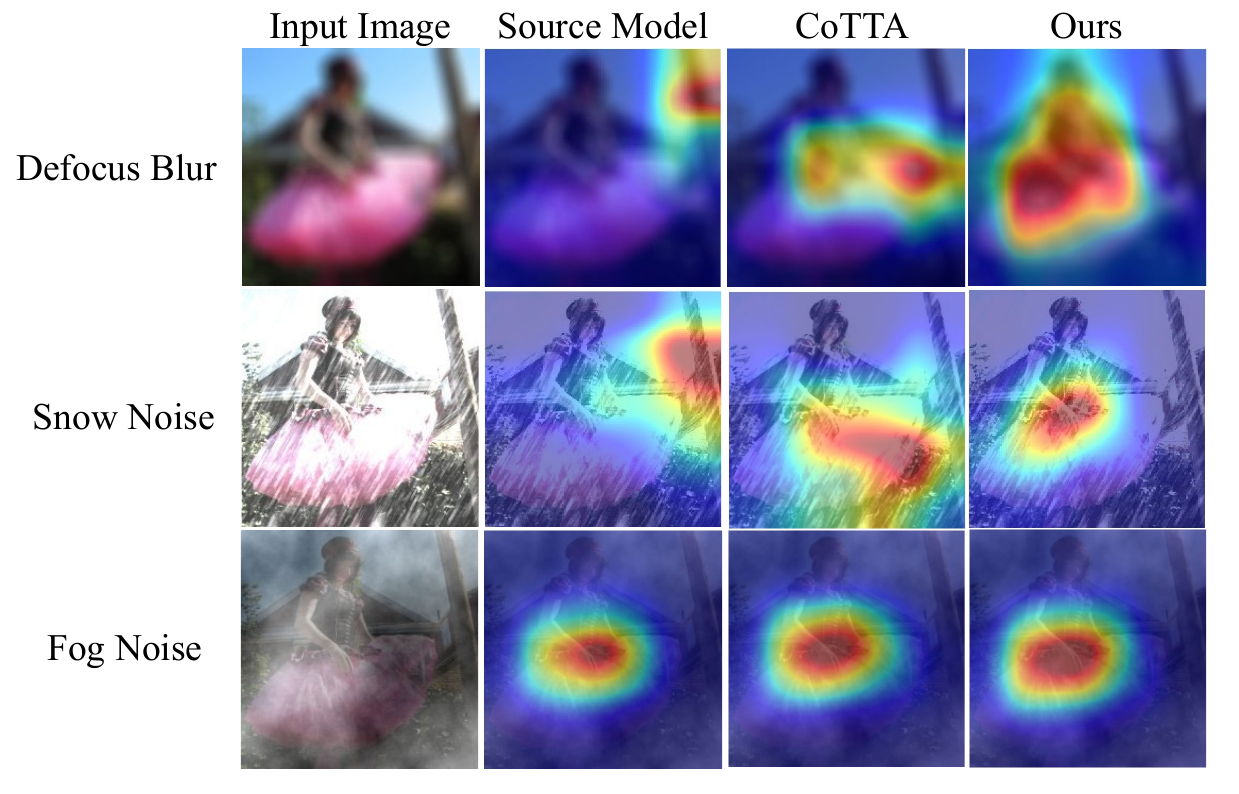}
  \caption{The CAM visualizations. These images belong to the \emph{hoopskirt} category.}
  \label{fig:cam}
\end{figure}

\begin{table}[tb]
\belowrulesep=0pt
\aboverulesep=0pt
\normalsize
\caption{Semantic segmentation results (mIoU in \%) on the Cityscapes-to-ACDC CTTA task. The four test conditions are repeated ten times to evaluate the long-term adaptation performance. All results are evaluated based on the Segformer-B5 architecture. Bold text indicates the best performance.}
\label{tab:segmentation_full}
\centering
\resizebox{0.98\textwidth}{!}{
\begin{tabular}{l|cccc|cccc|cccc|cccc|cccc|c} 
\toprule
Time & \multicolumn{20}{c|}{$t\xrightarrow{\makebox[\dimexpr 39\width][c]{\quad}}$} & ~ \\
\midrule
{Round} & \multicolumn{4}{l|}{1} & \multicolumn{4}{l|}{2} & \multicolumn{4}{l|}{3} & \multicolumn{4}{l|}{4} & \multicolumn{4}{l|}{5} & cont.  \\ 
\midrule
\multirow{3}*{Condition} & \multirow{3}*{\rotatebox{75}{Fog}} & \multirow{3}*{\rotatebox{75}{Night}} & \multirow{3}*{\rotatebox{75}{Rain}} & \multirow{3}*{\rotatebox{75}{Snow}} & \multirow{3}*{\rotatebox{75}{Fog}} & \multirow{3}*{\rotatebox{75}{Night}} & \multirow{3}*{\rotatebox{75}{Rain}} & \multirow{3}*{\rotatebox{75}{Snow}} & \multirow{3}*{\rotatebox{75}{Fog}} & \multirow{3}*{\rotatebox{75}{Night}} & \multirow{3}*{\rotatebox{75}{Rain}} & \multirow{3}*{\rotatebox{75}{Snow}} & \multirow{3}*{\rotatebox{75}{Fog}} & \multirow{3}*{\rotatebox{75}{Night}} &\multirow{3}*{\rotatebox{75}{Rain}} & \multirow{3}*{\rotatebox{75}{Snow}} & \multirow{3}*{\rotatebox{75}{Fog}} & \multirow{3}*{\rotatebox{75}{Night}} & \multirow{3}*{\rotatebox{75}{Rain}} & \multirow{3}*{\rotatebox{75}{Snow}} & \multirow{3}*{cont.} \\
~ & ~ & ~ & ~ & ~ & ~ & ~ & ~ & ~ & ~ & ~ & ~ & ~ & ~ & ~ & ~ & ~ & ~ & ~ & ~ & ~ & ~ \\
~ & ~ & ~ & ~ & ~ & ~ & ~ & ~ & ~ & ~ & ~ & ~ & ~ & ~ & ~ & ~ & ~ & ~ & ~ & ~ & ~ & ~ \\
\midrule
Source & 69.1 & 40.3 & 59.7 & 57.8 & 69.1 & 40.3 & 59.7 & 57.8 & 69.1 & 40.3 & 59.7 & 57.8 & 69.1 & 40.3 & 59.7 & 57.8 & 69.1 & 40.3 & 59.7 & 57.8 & cont. \\
BN Stats Adapt & 62.3 & 38.0 & 54.6 & 53.0 & 62.3 & 38.0 & 54.6 & 53.0 & 62.3 & 38.0 & 54.6 & 53.0 & 62.3 & 38.0 & 54.6 & 53.0 & 62.3 & 38.0 & 54.6 & 53.0 & cont. \\
TENT-cont.\cite{wang2021tent} & 69.0 & 40.2 & 60.1 & 57.3 & 68.3 & 39.0 & 60.1 & 56.3 & 67.5 & 37.8 & 59.6 & 55.0 & 66.5 & 36.3 & 58.7 & 54.0 & 65.7 & 35.1 & 57.7 & 53.0 & cont. \\
CoTTA\cite{Wang_2022_CVPR} & 70.9 & 41.2 & 62.4 & 59.7 & 70.9 & 41.1 & 62.6 & 59.7 & 70.9 & 41.0 & 62.7 & 59.7 & 70.9 & 41.0 & 62.7 & 59.7 & 70.9 & 41.0 & 62.8 & 59.7 & cont.\\
Ours & 71.2 & 42.3 & 64.9 & 62.0 & 72.6 & 43.2 & 66.3 & 63.2 & 72.8 & 43.8 & 66.5 & 63.2 & 72.8 & 43.6 & 66.7 & 63.3 & 72.6 & 43.5 & 66.5 & 63.0 & cont. \\

\midrule
{Round} & \multicolumn{4}{l|}{6} & \multicolumn{4}{l|}{7} & \multicolumn{4}{l|}{8} & \multicolumn{4}{l|}{9} & \multicolumn{4}{l|}{10} & {Mean$\uparrow$}  \\ 
\midrule
Source & 69.1 & 40.3 & 59.7 & 57.8 & 69.1 & 40.3 & 59.7 & 57.8 & 69.1 & 40.3 & 59.7 & 57.8 & 69.1 & 40.3 & 59.7 & 57.8 & 69.1 & 40.3 & 59.7 & 57.8 & 56.7 \\
BN Stats Adapt & 62.3 & 38.0 & 54.6 & 53.0 & 62.3 & 38.0 & 54.6 & 53.0 & 62.3 & 38.0 & 54.6 & 53.0 & 62.3 & 38.0 & 54.6 & 53.0 & 62.3 & 38.0 & 54.6 & 53.0 & 52.0 \\
TENT-cont.\cite{wang2021tent} & 64.9 & 34.0 & 56.5 & 52.0 & 64.2 & 32.8 & 55.3 & 50.9 & 63.3 & 31.6 & 54.0 & 49.8 & 62.5 & 30.6 & 52.9 & 48.8 & 61.8 & 29.8 & 51.9 & 47.8 & 52.3 \\
CoTTA\cite{Wang_2022_CVPR} & 70.9 & 41.0 & 62.8 & 59.7 & 70.9 & 41.0 & 62.8 & 59.7 & 70.9 & 41.0 & 62.8 & 59.7 & 70.8 & 41.0 & 62.8 & 59.7 & 70.8 & 41.0 & 62.8 & 59.7 & 58.6 \\


Ours & 72.5 & 43.0 & 66.5 & 62.8 & 72.5 & 42.5 & 66.8 & 63.3 & 72.5 & 43.1 & 66.9 & 63.0 & 72.5 & 43.5 & 67.1 & 63.2 & 72.5 & 42.9 & 66.7 & 63.0 & \textbf{61.3} \\
\bottomrule
\end{tabular}
}
\end{table}


\section{Additional Experiments on CAM}\label{cam}
To further substantiate the efficacy of our approach, we leverage CAM visualizations on the ImageNet-C dataset, offering a visual testament to our method's adaptability. 
As depicted in \cref{fig:cam}, when utilizing only the source model, significant differences in feature attention across different domains are observed. This indicates that the features extracted by the source model are highly susceptible to domain shifts, leading to difficulties in focusing on foreground objects. The strong comparison method CoTTA\cite{Wang_2022_CVPR} can appropriately adapt the model to continually changing target domains. However, this method still suffers from the issue of dispersed feature attention and a tendency to focus on simple background features in the online unsupervised data stream. In contrast, our method enables the output features to overlook changing background domain shifts and consistently highlight foreground objects with higher response values.

\section{Experiment Results on Cityscapes-to-ACDC}\label{seg_ctta}
In dense prediction semantic segmentation tasks, each pixel point can be approximated as a sample belonging to a certain category. Therefore, in the semantic segmentation CTTA task, considering memory constraints and timeliness, we perform uncertainty-aware buffering and organizing of pixels on the current input data. In other words, the buffer size can be considered as 1. We have implemented our method based on the previous state-of-the-art method CoTTA~\cite{Wang_2022_CVPR}.

We show the comprehensive experiment results on the Cityscapes-to-ACDC CTTA task in \cref{tab:segmentation_full}. It is evident that both the BN Stats Adapt method and the TENT-based method exhibit weaker performance compared to the source model on this complex task. Furthermore, the performance of the TENT-based method deteriorates significantly over time. In contrast, our method achieves an absolute improvement of 4.6\% compared to the source model without adaptation. Particularly noteworthy is the comparison with the previous state-of-the-art method CoTTA~\cite{Wang_2022_CVPR}, wherein our method demonstrates a clear trend of gradual improvement in the average mIoU metric during the initial rounds and consistently maintains high mIoU performance in the long-term adaptation. This showcases the ability of our method to buffer and organize reliable information embedded in the unsupervised data stream, thereby contributing to the adaptation progress.

\section{Additional Experiments on RoTTA}\label{ptta}
\begin{table}[tb]
\belowrulesep=0pt
\aboverulesep=0pt
\caption{Classification error rate (\%, lower is better) for different methods. All results are evaluated with the largest corruption severity
level 5 in an online manner. Bold text indicates the best performance.}
\label{tab:ptta}
\centering
\begin{tabular}{c|ccc|c} 
\toprule
Method & CIFAR10C & CIFAR100C & ImageNet-C & Mean$\downarrow$ \\
\midrule
CoTTA\cite{Wang_2022_CVPR} & 16.2 & 32.5 & 62.7 & 37.1 \\
RoTTA\cite{yuan2023robust} & 19.3 & 34.8 & 67.3 & 40.5 \\
Ours & \textbf{15.8} & \textbf{29.0} & \textbf{59.0} & \textbf{34.6} \\
\bottomrule
\end{tabular}
\end{table}


We note that there is recent work\cite{yuan2023robust} on utilizing a buffer to cope with temporal correlation and generate i.i.d. data streams in scenarios where the environments gradually change and the test data is sampled correlatively over time. Here, we apply this method under the CTTA setting and present the comparative results in \cref{{tab:ptta}}. Despite the use of a buffer mechanism, RoTTA\cite{yuan2023robust} is generally weaker than the CoTTA\cite{Wang_2022_CVPR} method across three CTTA datasets. In contrast, our method reshapes the online data buffering and organizing mechanism in CTTA, dynamically aggregating reliable information from the unsupervised online data stream through an uncertainty-aware buffering approach. Furthermore, our method mitigates error accumulation and catastrophic forgetting problems in CTTA through two well-designed constraints. Experimental results demonstrate the effectiveness of our buffering and organizing mechanism in simultaneously addressing three key challenges in CTTA, showing significant improvements over RoTTA across all three datasets.

\end{document}